\documentclass[10pt,twocolumn,letterpaper]{article}

\usepackage{iccv}
\usepackage{times}
\usepackage{epsfig}
\usepackage{graphicx}
\usepackage{amsmath}
\usepackage{amssymb}
\usepackage{booktabs}
\usepackage{cite}

\usepackage[pagebackref=true,breaklinks=true,letterpaper=true,colorlinks,bookmarks=false]{hyperref}

\iccvfinalcopy 

\def\iccvPaperID{4746} 

\ificcvfinal\fi

\def\mn{LCGU}

\begin{document}

\title{Looking into a Pixel by Nonlinear Unmixing - A Generative Approach} 

\author{Maofeng Tang$^1$, Hairong Qi$^1$\\
$^1$Department of Electrical Engineering and Computer Science, University of Tennessee, Knoxville\\
{\tt\small mtang4@vols.utk.edu,  hqi@utk.edu}
}

\maketitle
\ificcvfinal\thispagestyle{empty}\fi

\begin{abstract}
Due to the large footprint of pixels in remote sensing imagery, hyperspectral unmixing (HU) has become an important and necessary procedure in hyperspectral image analysis. Traditional HU methods rely on a prior spectral mixing model, especially for nonlinear mixtures, which has largely limited the performance as well as generalization capacity of the unmixing approach. In this paper, we exploit the challenging problem of hyperspectral nonlinear unmixing (HNU) \textit{without} explicitly knowing the mixing model. 
Inspired by the principle of generative models where images of the same distribution can be generated as that of the
training images without knowing the exact probability distribution function of the image, we develop an invertible
mixing-unmixing process via a bi-directional GAN frame-
work, constrained by both the cycle consistency and the linkage between linear and nonlinear mixtures. The combination of cycle consistency and linear linkage provides such powerful constraints and without the need of an explicit mixing model.  We refer to the proposed approach as linearly-constraint CycleGAN unmixing net, or \mn~ net. Experimental results indicate that the proposed \mn~ net exhibits stable and competitive performance on different datasets as compared to other state-of-the-art HNU methods that are model-based.
\end{abstract}

\section{Introduction}
\label{Intro}

Remote sensing imagery has been increasingly utilized in recent decades for large-scale Earth surface monitoring, where hyperspectral images  are among the most popularly used. Compared to conventional RGB 
images with high spatial resolution, hyperspectral images collect hundreds of contiguous bands which provide finer spectral details but also poor spatial resolution due to hardware limitations. This has resulted in the so-called ``mixed pixels'', where more
than one type of material is present within a single pixel due to its large footprint. Thus, the measured
spectrum of a single pixel is a mixture of several ground cover
spectra known as \textit{endmembers}, weighted by their fractional
\textit{abundances}. Thus  hyperspectral unmixing (HU) that refers to the procedure of deriving the 
endmembers and their corresponding  abundances, plays a fundamental role in hyperspectral image analysis and has contributed significantly in a series of downstream tasks, such as classification, segmentation and object detection~\cite{gross2021multi, vibhute2021hyperspectral}.

The classical HU methods can be categorized into two groups according to the mixing models used,~\ie, linear mixing model (LMM) and nonlinear mixing model (NLMM). The LMM has been broadly adopted in a wide range of applications due to its simplicity and intuitive physical interpretation. However, it makes two general assumptions that might not always hold in real-world applications: 1) the mixing scale is macroscopic, 
\ie, the materials are mixed homogeneously~\cite{singer1979mars}, and 2) the photons that reach the sensor interact with only one material\cite{clark1984reflectance}. 
In reality, endmembers tend to interact with each other under complex scenarios \cite{bioucas2012hyperspectral}. 
For example, 
there are common situations when interactions occur at a heterogeneous region, 
yielding the so-called  \textit{intimate mixture}\cite{hapke2012theory}. Such mixtures have been observed for scenes composed of sand or mineral mixtures \cite{nash1974spectral}. 
Another type of nonlinear mixture occurs when there exist multiple interactions among scattered light at different layers, 
yielding the so-called \textit{multilayered mixture}. This is often the case for images acquired over forested areas, where there may be many interactions between the ground and the canopy of forest.

To address the challenges above, 
increasing attention has been paid to the study of nonlinear unmixing. 
For example, 
there have been investigations done by analyzing intimate mixtures observed 
in laboratory \cite{mustard1989photometric}. The Hapke model \cite{hapke2012theory}, developed based on the Radiation Transmission (RT) theory, is another popular approach dealing with the intimate mixtures. 
On the other hand, 
several models \cite{dobigeon2013nonlinear} have been proposed to analytically describe the multilayered interactions, which usually consist of the power of the products of the reflectance. For instance, the bilinear mixture models \cite{halimi2011nonlinear} describe the nonlinear mixtures from interactions of only two endmembers in a pixel. This was later extended to involve multiple endmembers, such as the p-Linear mixture model \cite{marinoni2015novel}.  
To simulate all possible reflections among endmembers, the multilinear mixing model (MLM) \cite{heylen2015multilinear} was proposed, which extends the polynomial post-nonlinear model (PPNM) \cite{altmann2012supervised} to infinite orders. Recently, the multi-harmonic post-nonlinear mixture model \cite{tang2018multiharmonic}, an extension of MLM, was proposed, which incorporates both the local high-order mixture and multiple reflections. 

These state-of-the-art hyperspectral nonlinear unmixing (HNU) approaches, although different, share a common mechanism that they all assume an explicit mixing 
model parameterized with abundance, and thus are all \textit{model-based}. However, there are two potential issues in model-based unmixing,~\ie, model generalization and model selection. The ``generalization'' issue refers to the fact that the models are difficult to be generalized to different regions. Since the models are
usually specifically designed according to the types of spectral mixtures in certain observed regions, it is highly unlikely that one model performs superior on the observed region would perform as effectively on a different region. 
The ``selection'' issue refers to the fact that 
in real applications, very often, multiple types of mixtures may exist in one observed region, making it difficult to select an appropriate model to fit the entire region, especially when there is no prior information about that region. 

To overcome these issues, it is imperative to propose   \textit{model-free or data-driven} methods, where the mixture model itself is learnable and flexible enough to reflect the corresponding mixtures in different regions, according to the characteristics of the data itself instead of a fixed mixing model. 

How can we perform robust nonlinear unmixing without having to explicitly derive the mixing model? We turn to generative adversarial networks (GANs) \cite{goodfellow2020generative}, one of the most important milestones in the evolution of deep learning. The vanilla GANs have been developed as a framework to generate images of the same distribution as that of the training images without knowing the exact probability distribution function of the image, thus solving a long-lasting problem in image generation. 
%
Here, we ask the same question: can we ``learn'' the  unmixing model without assuming the knowledge of the nonlinear mixing model? 


The main contributions of this paper can be summarized as follows:
First, it introduces the GAN framework to the HNU problem, enabling the truly data-driven (or model-free) nonlinear unmixing.   
Second, it realizes an invertible mixing-unmixing process via a bi-directional GAN framework. The bi-directional network derives the mixture model from two data flows, including the  mixing-unmixing and the unmixing-mixing flows, which makes the learned mixture model stronger and more reliable.
Third, it exploits the intrinsic relationship between the nonlinear mixture and linear mixture and uses it as a constraint to stablize the solution. 


\section{Related Work}
\label{relate work}


Although deep learning, as a powerful nonlinear representation learning framework, has been in development since 2006 \cite{hinton2006reducing}, its introduction to the field of spectral unmixing is much later \cite{guo2015hyperspectral}, where an autoencoder (AE) cascade structure was proposed to solve the linear unmixing problem. Since then, the AE structure has been heavily studied for hyperspectral unmixing. Several attempts  \cite{palsson2018hyperspectral,qu2018udas,su2019daen,palsson2020convolutional,hong2021endmember,gao2021cycu} were made to use AE based deep learning structures for unsupervised linear unmixing, where the output from the encoder serves as the abundance vector and the weight vectors of the decoder serve as the endmembers.  
With these models, the hyperspectral image would be reconstructed using the linear combination of the weight vectors in one layer of the decoder. However, these AE-based methods are with two obvious shortages: (1) The model is built under the linear mixture assumption, which lacks the intrinsic mechanism to realize the nonlinear mixture which is more common in real world; (2) Because the training process is unsupervised without imposing any constraints, to fulfill the physical meaning of abundance and endmeber, sum to one and non negative, hard coding is necessary in these methods, such as normalization. 

To fix the nonlinear mixture, more learning-based models have been proposed. \cite{koirala2019supervised} proposed to address the nonlinear unmixing problem using existing nonlinear mixture models to generate the synthetic data as the training set. Su~\etal~\cite{su2020deep} developed an unsupervised multitask learning-based method to address the problem of nonlinear unmixing. However, it is specifically designed to derive the abundance from the bilinear mixtures. More recently, in \cite{wang2019nonlinear}, a postnonlinear decoder is designed by adding nonlinear layers to a linear decoder. In \cite{zhao2021lstm, zhao2021hyperspectral}, instead of adding subsequential nonlinear layers, a network branch is added in parallel to the linear layer for constructing additive nonlinear structures. Although these models have capability to simulate nonlinear mixture, they still hard to represent the physical constraint of abundance and endmember and less explainable for them. To summarize, existing deep learning-based unmixing methods generally have two drawbacks: (1) for those totally model-free methods, the correct training set is necessary but hard to collect, and these methods are hard to use in the large scale image with variant spectral feature; (2) for those methods without abundance ground truth, the prior mixture model is necessary, but the bias between real spectral mixture and prior model always exist and hard to eliminate.

\begin{figure*}[t]
  \centering
  \includegraphics[width=6.8in]{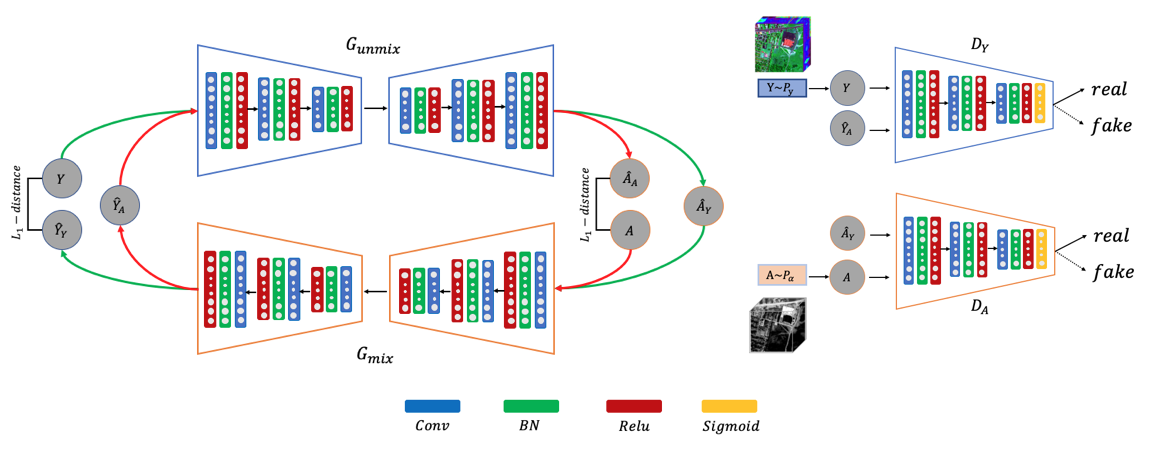}\\
  \caption{The cycle consistency in unmixing. Left: Summary of the bi-directional data flow. Right-top: the unmixing-mixing flow in the raw image domain. Right-bottom: the mixing-unmixing flow in abundance domain.}\label{fig:cyclegan}
  \vspace{-5mm}
\end{figure*}

\section{Proposed Method: \mn}\label{method}
We address the problem of model-free hyperspectral nonlinear unmixing through a data-driven linearly-constrained CycleGAN framework, referred to as LCGU. 

\subsection{Nonlinear Unmixing as a Generative Learning Problem}
\label{sec:num-gen}

Inspired by GAN where images with the same distribution as that of the training images can be generated without knowing the explicit expression of the distribution itself, here we ask the same question if the abundance maps/images can be generated without knowing explicitly the mixing or unmixing function. 
Our initial trial of directly applying GAN or its variants actually failed. This is because to generate the correct abundance maps for a given image with GAN, it requires a large amount of ground truth abundance maps, which is impractical in remote sensing applications. 

From the perspective of generative learning, the procedure of deriving the abundance from the raw image is essentially the same as finding a set of low-dimensional representations (abundance) that can reconstruct the raw image well~\cite{qu2018udas}. We denote the domain of the raw hyperspectral image, $\mathcal{Y}$, as the source domain and the domain of the abundance image, $\mathcal{A}$, as the target domain. The unmixing procedure transfers the image from the source domain $\mathcal{Y}$ to the target domain $\mathcal{A}$, and the mixing procedure transfers the image from the target domain $\mathcal{A}$ to the source domain $\mathcal{Y}$, as shown in Fig.~\ref{fig:cyclegan}. Such a mixing-unmixing transformation can be interpreted as an image-to-image transformation problem, assuming the endmembers are known. 

To generate images with the correct distributions in different domains, \ie, the abundance map $\mathbf{A}$ in the target domain, and the raw images $\mathbf{Y}$ in the source domain, 
we need to construct two discriminators, $D_{A}$ and $D_{Y}$, in these two domains, respectively, as shown in Fig.~\ref{fig:cyclegan}. For the unmixing procedure, since we don't have the ground truth abundance $\mathbf{A}$, the discriminator $D_{A}$ is adopted to enforce the generated abundance map to follow a Dirichlet distribution \cite{qu2018unsupervised}, which naturally satisfies the abundance sum-to-one and the abundance nonnegative physical constraints. In addition, 
we introduce the so-called cycle consistency constraint from CycleGAN to further regulate the solution space, where the mixing $G_{mix}$ and unmixing $G_{unmix}$ procedures are both reversible in a bi-directional data flow. More details will be described in Sec.~\ref{sec:bi}. To preserve the semantic information of the derived abundance, the abundance is further regularized based on the semantic similarity between the linear mixture and the raw non-linear mixing data. This part will be elaborated in Sec.~\ref{sec:linear}.

\subsection{Cycle Consistency through Bi-Directional Data Flow}
\label{sec:bi}

The data flow of the bi-directional structure includes two branches, as shown in the right part of Fig.~\ref{fig:cyclegan}, \ie, the unmixing-mixing branch ${\mathcal{Y}{\rightarrow}\mathcal{A}{\rightarrow}\mathcal{Y}}$ and the mixing-unmixing branch $\mathcal{A}{\rightarrow}\mathcal{Y}{\rightarrow}\mathcal{A}$. The unmixing-mixing branch learns the abundance $\hat{\mathbf{A}}$ from the image $\mathbf{Y}$ with $G_{unmix}$, as shown on the right top of Fig.~\ref{fig:cyclegan}. Instead of adopting a fixed mixing model, we first concatenate the estimated abundance with the 
known endmember  $\mathbf{M}$, and then learn a mapping function from the stacked data to the raw nonlinear mixing data with $G_{mix}$, such that the abundance can be derived without knowing the nonlinear mixing model explicitly. The mixing-unmixing branch shares the same weights as the unmixing-mixing branch. It projects the abundance $\mathbf{A}$ from the abundance domain to the image domain with $G_{mix}$, and then reconstructs the abundance $\hat{\mathbf{A}}$ from $\hat{\mathbf{Y}}$ with $G_{unmix}$, as shown at the right-bottom of Fig.~\ref{fig:cyclegan}. Two discriminators $D_A$ and $D_Y$ are constructed in the middle of the two branches to enforce $\hat{\mathbf{A}}$ and $\hat{\mathbf{Y}}$ to follow the distributions of the abundance and the raw image with GAN loss, as described in Eqs.~\eqref{eq:ganA} and \eqref{eq:ganY}, respectively. Note that, since the ground truth of the abundance map is unavailable, the $\mathbf{A}$ as input to the discriminator $D_A$ in the unmixing-mixing branch and the $\mathbf{A}$ as input to the mixing-unmixing branch are drawn from the Dirichlet distribution. 


\begin{equation}\label{eq:ganA}
    \begin{aligned}
     \mathcal{L}_{GAN}(G_{unmix}&, D_A, \mathbf{Y}, \mathbf{A}) = 
     E_{A \sim p_{data(A)}}[{\rm{log}} D_A(A)] \\
     &\quad + E_{y \sim p_{data(Y)}}[{\rm{log}}(1 - D_A(G_{unmix}(y)))].
    \end{aligned}
\end{equation}

\begin{equation}\label{eq:ganY}
    \begin{aligned}
     \mathcal{L}_{GAN}(G_{mix}, & D_Y, A, Y) = {E_{Y{\sim}p_{data(\mathcal{Y})}}[{\rm{log}}D_Y(Y)}] \\
  & + {E_{A{\sim}p_{data(\mathcal{A})}}[{\rm{log}}(1-D_Y(G_{mix}(A)))}]
    \end{aligned}
\end{equation}


We summarize the data flow of the bi-directional unmixing network in Fig.~\ref{fig:cyclegan}. In the ideal case, given an individual input $\mathbf{Y}$ in the source domain $\mathcal{Y}$, the $G_{unmix}$ function should map it to the desired abundance $\hat{\mathbf{A}}$ in the target domain $\mathcal{A}$. However, the current structure cannot guarantee that the generated $\hat{\mathbf{A}}$ and the input $\mathbf{Y}$ are paired up. That is, with only the GAN loss, the function $G_{unmix}$ can map an input $\mathbf{Y}$ to multiple images that correspond to the distribution of the target domain. Therefore, to further regularize the solution space, we introduce the cycle-consistency loss. That is, given an input image $\mathbf{Y}$, the unmixing-mixing branch/cycle is able to transfer the input back to the original image $\mathbf{Y}{\rightarrow}G_{unmix}({\mathbf{Y}}){\rightarrow}G_{mix}(G_{unmix}({\mathbf{Y}}))\approx{\mathbf{Y}}$. Similarly, the mixing-unmixing cycle should satisfy the criteria $\mathbf{A}{\rightarrow}G_{mix}({\mathbf{A}}){\rightarrow}G_{unmix}(G_{mix}({\mathbf{A}}))\approx{\mathbf{A}}$. Based on these two cycles, the cycle-consistency loss is defined as 




\begin{equation}\label{eq:cycle}
\begin{aligned}
\mathcal{L}_{re}(&G_{unmix}, G_{mix}) = \\
& \, E_{\mathbf{Y} \sim p_{data}(\mathcal{Y})} \big[ ||G_{mix}(G_{unmix}(\mathbf{Y})) - \mathbf{Y}||_1 \big] \\
& + E_{\mathbf{A} \sim p_{data}(\mathcal{A})} \big[ ||G_{unmix}(G_{mix}(\mathbf{A})) - \mathbf{A}||_1 \big].
\end{aligned}
\end{equation}

It is worth mentioning that, compared to the uni-directional reconstruction loss used in the autoencoder-based unmixing approaches, the cycle-consistency loss is able to regularize the solution space without requiring the mixing model to be explicitly defined. This would largely facilitate the bi-directional network to derive the desired abundance based on the characteristics of the input data.


\subsection{Semantic Consistency between Linear and Nonlinear Formulations}
\label{sec:linear}
In both the unmixing-mixing and mixing-unmixing cycles, the abundance $\mathbf{A}$ is concatenated with the endmember $\mathbf{M}$, and $\mathbf{A}$ is drawn randomly from the Dirichlet distribution. This randomness would very likely cause unstable abundance estimation, leading to the reconstructed raw image being unstable. To further regularize the solution space, we introduce another constraint, referred to as the ``semantic consistency'' constraint to preserve the semantic information of the reconstructed image as well as the estimated abundance map. The rationale is that although the abundance map estimated by the nonlinear unmixing is different from that by linear unmixing,  they should still hold similar semantic structures, since they are both derived from the same raw image based on the same endmembers. Therefore, the linear combination of the endmembers using the abundance derived from the nonlinear unmixing should be similar to the raw data. 
This hypothesis has been shown to be effective in \cite{koirala2019supervised}.  

Hence, in the proposed method, besides the cycle-consistency constraint in the bi-directional unmixing network, we also incorporate the semantic consistency as an additional constraint to help preserve the semantic information of the derived abundance. This is realized using a pre-trained autoencoder $AE_p$, as shown in the bottom of Fig.~\ref{fig:cyclegan}. 
In the pre-training stage, the input of the autoencoder $AE_p$ is the raw image $\mathbf{Y}$, and the network is trained to minimize the distance between $\mathbf{Y}$ and the reconstructed image $\hat{\mathbf{Y}}$. 
During the nonlinear unmixing procedure, the input of $AE_p$ is  the linear combination of the abundance map estimated from the bi-directional nonlinear unmixing and the endmembers $\hat{\mathbf{A}}_Y \times \mathbf{M}$ and the output of the network is the reconstructed image $\hat{Y}_{AM}$. 
Since $\hat{\mathbf{A}}_Y \times \mathbf{M}$ should preserve the semantic information of  the raw image and the network $AE_p$ was pre-trained to reconstruct the original raw image, 
the reconstruction loss between $\hat{\mathbf{A}}_Y \times \mathbf{M}$ and $\hat{Y}_{AM}$, as $loss_3$, should be minimized. 
In addition, the mutual information $loss_4$ between the reconstructed image $\hat{Y}_{AM}$ and the raw image $\mathbf{Y}$ is introduced to further enforce the semantic similarity. It is worth mentioning that here we adopt the mutual information instead of the reconstruction loss to preserve the global semantic information of the abundance instead of local similarities. The $loss_3$ and $loss_4$ are defined in Eqs.~\eqref{eq8} and $\eqref{eq9}$,

\begin{equation}\label{eq8}
{\mathcal{L}_{{AE_p}-RE} = {E_{A{\sim}G_{unmix}}[||AE_p-\hat{\mathbf{A}}_Y\times \mathbf{M})||_1]}}
\end{equation}
\begin{equation}\label{eq9}
{\mathcal{L}_{{AE_p}-MI} = {E_{A{\sim}G_{unmix}}[\mathcal{MI}(AE_P, \mathbf{Y})]}}
\end{equation}
with
\begin{align}
\mathcal{MI}\left(AE_P ; \mathbf{Y}\right) & = H\left(AE_P\right)-H\left(AE_P \mid \mathbf{Y}\right)\\ & = \int_{AE_P \times \mathbf{Y}} \log \frac{\mathbb{P}_{AE_P \mathbf{Y}}}{\mathbb{P}_{AE_P} \otimes \mathbb{P}_{\mathbf{Y}}} d \mathbb{P}_{AE_P \mathbf{Y}},
\end{align}

where $H$ indicates the Shannon entropy and $H\left(AE_P \mid \mathbf{Y}\right)$ is the conditional entropy of $AE_P$ given $\mathbf{Y}$. $AE_P$ is the output of pre-trained auto encoder with the input $\hat{\mathbf{A}}_Y\times \mathbf{M}$. The exact mutual information is difficult to calculate, so we resort to an estimation of it in two local patches based on the MINE algorithm, which is implemented by back propagation in two layers fully connection network\cite{belghazi2018mutual}.

\subsection{Optimization and Implementation Details}
Based on Eqs.~\ref{eq:ganA} to \ref{eq9}, the objective function of the proposed \mn~ can then be expressed as:
\begin{equation}\label{eq11}
    \begin{aligned}
     \mathcal{L_{\mn}} = &\mathcal{L}_{GAN}(G_{unmix}, D_A, Y, A)\\
     &+\mathcal{L}_{GAN}(G_{mix}, D_Y, \hat{A}, Y)\\
     &+ \mathcal{L}_{Re}(G_{unmix}, G_{mix})\\
     &+\mathcal{L}_{Re}(G_{mix}, G_{unmix})\\
     &+ \mathcal{L}_{{AE_p}-RE} + \mathcal{L}_{{AE_p}-MI}\\
    \end{aligned}
\end{equation}

To find the optimal solution to Eq.~\ref{eq11}, we highlight the following implementation details.
First of all, most deep-learning-based spectral unmixing approaches would process the image through a pixel-wise training procedure, which did not effectively incorporate the spatial correlation within a local neighborhood.  To address this problem, instead of pixel-wise input, we divide the raw image into $32\times 32\times L$ patches with $1/3$ overlapping areas, and feed the patches to the network for training purpose, 
such that both the spectral and spatial characteristics can be considered during the unmixing process. 
The pixel values are normalized between 0 and 1 before training. 

Both the unmixing $G_{unmix}$ and mixing networks $G_{mix}$ are constructed with a 5-layer Conv-Deconv structure. 
The discriminators, $D_Y$'s and $D_A$'s are constructed with 3 convolutional layers, and a fully-connected layer with a sigmoid activation function as the output layer. In addition, for the input of $G_{mix}$, the abundance map $\hat{\mathbf{A}}_Y$ is concatenated to the resized endmembers, so the number of channels is $L+R$. 

In the training procedure, we adopt ADAM ($\alpha = 0.0002$, $\beta = 0.5$) as the optimization method and train the network with 25 epochs.

\section{Experiments and Analysis}\label{experiment}
\subsection{Experimental Design and Performance Metrics}
The proposed \mn~is evaluated via the synthetic data from three perspectives: the effectiveness of the algorithm, the generalization capacity of the algorithm and the necessity of the major components of the network structure (or ablation study). 

In the first set of experiments, the synthetic images are generated with both linear and different nonlinear mixing models to simulate the complex mixing scenes in real hyperspectal images. Specifically, the mixing models include linear mixture model (LMM) \cite{bioucas2012hyperspectral}, bilinear mixing model (BMM) \cite{halimi2011nonlinear}, post-nonlinear mixing model (PNMM) \cite{altmann2012supervised}, and multilinear mixing model (MLM) \cite{heylen2015multilinear}, among which, both the BMM and PNMM are  2nd-order models covering different mixture types, and MLM is a high-order model (infinite). These models can, in general, comprehensively cover different types of mixtures in real scenery. The proposed \mn~is compared with both model-based and deep learning-based unmixing methods. The unmixing results are evaluated using two quantitative metrics of the abundance angle distance (AAD) and the abundance information divergence (AID). The lower the value of the metrics, the better the performance of the algorithm. 
 
The second set of experiments are designed to evaluate the generalization capacity of the proposed \mn~using synthetic data generated from three mixture models, \ie, LMM, PNMM and MLM. 
The proposed \mn~is trained on the data generated with a single mixing model, \eg, LMM, and tested on the data generated with other mixing models, \eg, PNMM. 
In addition, the proposed \mn~is also trained on the augmented dataset covering two mixing models, \eg, LMM and PNMM, and tested on the other dataset generated from a mixing model unseen in the training stage, \eg, MLM. 

The third set of experiments focus on the ablation study to evaluate three major components of the proposed network structure, \ie, the effectiveness of  1) the implicit supervision, 2) the bi-directions structure, and the mutual information based loss function between the linear mixture and nonlinear mixture. 
 
For the real data experiment, the comparison is conducted on both model-based and learning-based methods including FCLS, PPNM, MLM, uDAS and NN-LM. Due to the lack of the ground truth abundance, we choose the reconstruction error and spectral angle distance (SAD) as the performance metrics. The estimated abundance maps for each endmember are also visualized inspected and compared. 

\subsection{Data}

\subsubsection{Data Set Description}\label{datadescriptor}
The synthetic images of size 250 × 250 × 420 are generated using both  the linear mixing model and different nonlinear mixing models, as discussed in part V.A, with five spectral signatures (endmembers) following the rule used in \cite{tang2017integrating}. The five endmembers with 420 bands are selected from the USGS spectral library \cite{swayze1993us}. The selected endmembers cover the spectral reflections of five materials including Alunite, Calcite, Epidote, Kaolinite, and Buddingtonite. In order to simulate a hyperspectral scene with spatial correlations which is the inherent characteristics of remote sensing imagery, the generation of the synthetic image follows the procedure in \cite{chen2013nonlinear}. 

We also evaluate the proposed method on two real hyperspectral images. 
The first one is the Urban hyperspectral image with resolution $307 \times 307 \times 162$, and a spectral wavelength ranging from 400 nm  to 2500 nm. The spectral reflectance of Asphalt, Grass, Tree, and Roof are chosen as the endmembers in the experiments. The second real image dataset was collected by the Hyperspectral Digital Imagery Collection Experiment (HYDICE) sensor over Washington, D.C. (WDC), and a subset of $150 \times 150$ pixels was cropped from the original image for this experiment. The subset has 191 bands covering the wavelength range of 400 nm to 2500 nm, after removing low-SNR and water-vapor absorption bands. We choose 5 materials, , as endmembers in the experiments. 




\subsection{Results and Analysis of the Synthetic Data}\label{AA}

\begin{table*}
  \centering
  \caption{Abundance Angle Distance (AAD) on the Synthetic Data with Gaussian White Noise from SNR = 30 to 15 db}\label{tab:aad}
  \vspace{2mm}
  \setlength{\tabcolsep}{1.8mm}{
  \begin{tabular}{c|cccc|cccc|cccc}
  \toprule
  \multicolumn{1}{c|}{SNR(db)}&\multicolumn{4}{c|}{30}&\multicolumn{4}{c|}{20}&\multicolumn{4}{c}{15}\\
  \midrule
   {}&LMM&BMM&PNMM&MLM&LMM&BMM&PNMM&MLM&LMM&BMM&PNMM&MLM\\
    \midrule
    FCLS&0.0768&0.8369&1.0158&1.8135&0.0436&0.7421&1.0582&2.3155&0.3663&0.9200&0.9797&2.6703\\
    \midrule
    GBM&0.6387&0.1882&0.1801&1.3490&0.6325&0.2639&0.3114&0.9904 &0.5990&0.6333&0.7598&1.7018\\
    \midrule
    PPNM&0.0834&0.1518&0.0801&1.1704&0.2301&0.2546&0.1089& 0.8637&0.3956&0.6201&0.4531&1.4290\\  \midrule
    \midrule
    uDAS&0.0542&0.7194&0.9377&1.0426&0.0361&0.6011&1.0021&1.4900 &0.2752&0.8817&0.8933&1.9003\\
    \midrule
    NN-LM&0.0911&0.2132&0.1954&0.2818&0.3319&0.4457&0.4334&0.4671&0.4704&0.5871&0.6221&0.6901\\
    \midrule \midrule
    \mn&\textbf{0.0423}&\textbf{0.0547}&\textbf{0.0539}&\textbf{0.0919}&\textbf{0.0637}&\textbf{0.0884}&\textbf{0.0888}&\textbf{0.1095}&\textbf{0.1002}&\textbf{0.1233}&\textbf{0.1210}&\textbf{0.1353}\\
    \bottomrule
  \end{tabular}}
  \vspace{-3mm}
\end{table*}


\subsubsection{Effectiveness of the Proposed \mn}
Table~\ref{tab:aad} shows the AAD of the estimated abundance from both the model-based and learning-based methods. For the results from the model-based methods, \ie, FCLS, GBM and PPNM, we observe that an unmixing method can achieve a lower AAD if it is applied to the data generated with a matching mixing model, 
\ie, FCLS on LMM, GBM on BMM, and PPNM on PNMM. 
However, when the methods are applied to the data generated with unmatched mixing models, \eg, FCLS on BMM, the AAD would increase significantly. This phenomenon reveals and validates that the performance of model-based unmixing approaches relies much on the correctness of the mixing models assumed for the data. An inappropriate mixing model would largely deteriorate the performance of spectral unmixing.  

The results from the learning-based approaches show similar phenomenon, since they are also developed according to the pre-defined mixing model. The uDAS network is constructed based on the LMM, while NN-LM is trained based on different nonlinear unmixing models. 
We can observe that uDAS performs well on the linear mixing data,  but poorly on the nonlinear mixing models, which shows that the mixing model assumption have a non-negligible impact on the performance of unmixing performance. 
NN-LM shows competitive performance when it is trained on the synthetic data generated with LMM, BMM, and PNMM models, and tested on the matching datasets, \ie, LMM, BMM, and PNMM datasets, respectively. However, we can observe that when the trained model is tested on a dataset generated with a different mixing model, \ie, MLM,  
the AAD of NN-LM significantly increases, which reveals the strong dependency of the method with explicit mixing model assumptions. From another perspective, compared with the model-based methods and the linear-assumption learning-based method, the performance drop is small. This suggests that loose mixture model assumption and resort to the data-driven would benefit the generalization of the method. Compared with other methods, the proposed \mn~performs well and stable consistently regardless of the type of datasets tested. It demonstrates the effectiveness of the proposed \mn~on the hyperspectral images with implicit mixing models. In addition, the experimental results on the datasets with different SNRs also indicate that the proposed method is more robust to noise compared with the model-based or learning-based methods. 

The above observations are further supported by the quantitative study with AID, you can find the table in supplement.


\subsubsection{Generalization Capacity of the Proposed \mn}
We evaluate the generalization capacity of the proposed \mn~by performing training and testing on datasets generated with different mixing models. The results are shown in supplement. The left, middle and right columns show the evaluation results on the synthetic data with Gaussian noise of 30db, 20db, and 15db, respectively. 

The top rowof figures show the evaluation results when the model is trained on a dataset generated with a single mixing model and tested on the datasets with other mixing models. We can observe that, the proposed \mn~can get the best performance when the mixing model of the training dataset and testing dataset are matched. 
When the mixing model of the training and testing datasts are mis-matched, the AAD's for all unmixing methods increase as expected. 
However, compared with the results shown in Table~\ref{tab:aad}, it is important to point out that the results achieved by the proposed \mn~are better than the model-based unmixing methods or autoencoder-based unmixing method assurming the nonlinear mixture, and comparable to the NN-LM method which is trained and tested on the datasets covering the same mixing models. Specifically, observing the PNMM and MLM columns in each SNR set, the AAD of mismatching unmixing method is larger than \mn~with the mismatching training/testing datasets. For example, in 30db set, the AAD of FCLS and uDAS for PNMM mixture(1.0158 and 0.9377) is obvious larger than the \mn~trained on the LMM and tested on PNMM(0.1994). This suggests that the proposed \mn~can effectively learn the implicit mixing model from the training data under model-free assumption, which demonstrates the superior generalization capacity of the proposed method. At another point, the robustness, we can find that the testing performance of the proposed model trained on the single mixing model data varying in a smaller range in different SNR setting compared to other methods shown in the Table~\ref{tab:aad}. Specifically, in top row of the generative results figure, the biggest range of varying  of the AAD in different SNR is  0.1655(difference of AAD in 15db and 30db(15db/30db) for \mn~trained on LMM and testing on MLM) and the smallest one is 0.0193(difference of AAD in 15db and 30db(15db/30db) for \mn~trained on PNMM and testing on LMM). For comparison, we can observe the noise effect in Table~\ref{tab:aad} and can find that, for the matching mixture and unmxing method, the AAD difference between the different SNR setting is varing from 0.2895 for 15db/30db with FCLS and LMM mixture to 0.373 for 15db/30db with PPNM and PNMM mixture, which is obvious larger than the varying of \mn~ in different SNR setting. It indicates that the robustness of \mn~. 

We design another experiment with the training data generated using multiple mixing models, and the results are shown at the right column of the generative results figure. We can observe that the unmixing results have similar trends as compared to the previous experiments with single mixing data. 
Nonetheless, we make two important observations. First, the multi-model training strategy increases the generalization capacity of the proposed \mn~when the training and testing data consist of different types of mixing models. 
For example, 
when the model is trained on datasets covering both LMM and PNMM mixtures and tested on a MLM mixing dataset, its performance increases as compared to the one trained on a dataset covering a single mixture, \eg, LMM or PNMM, and tested on a MLM mixing dataset. 
\begin{figure}
  \centering
  \includegraphics[width=3.3in]{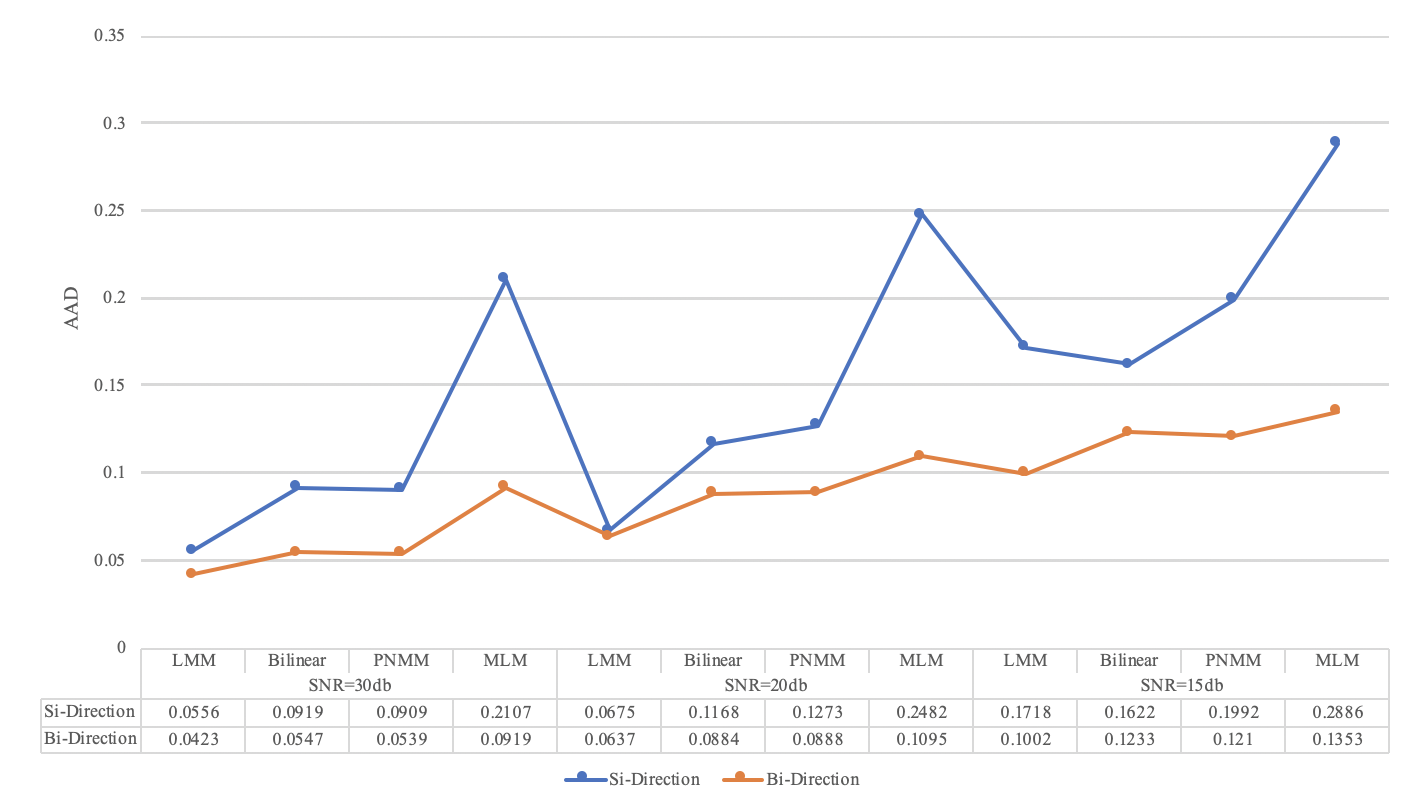}\\
  \caption{The effect of the bi-directional structure as compared to the uni-directional structure.}\label{fig:Bi}
  \vspace{-3mm}
\end{figure}

\begin{figure}
  \centering
  \includegraphics[width=3.3in]{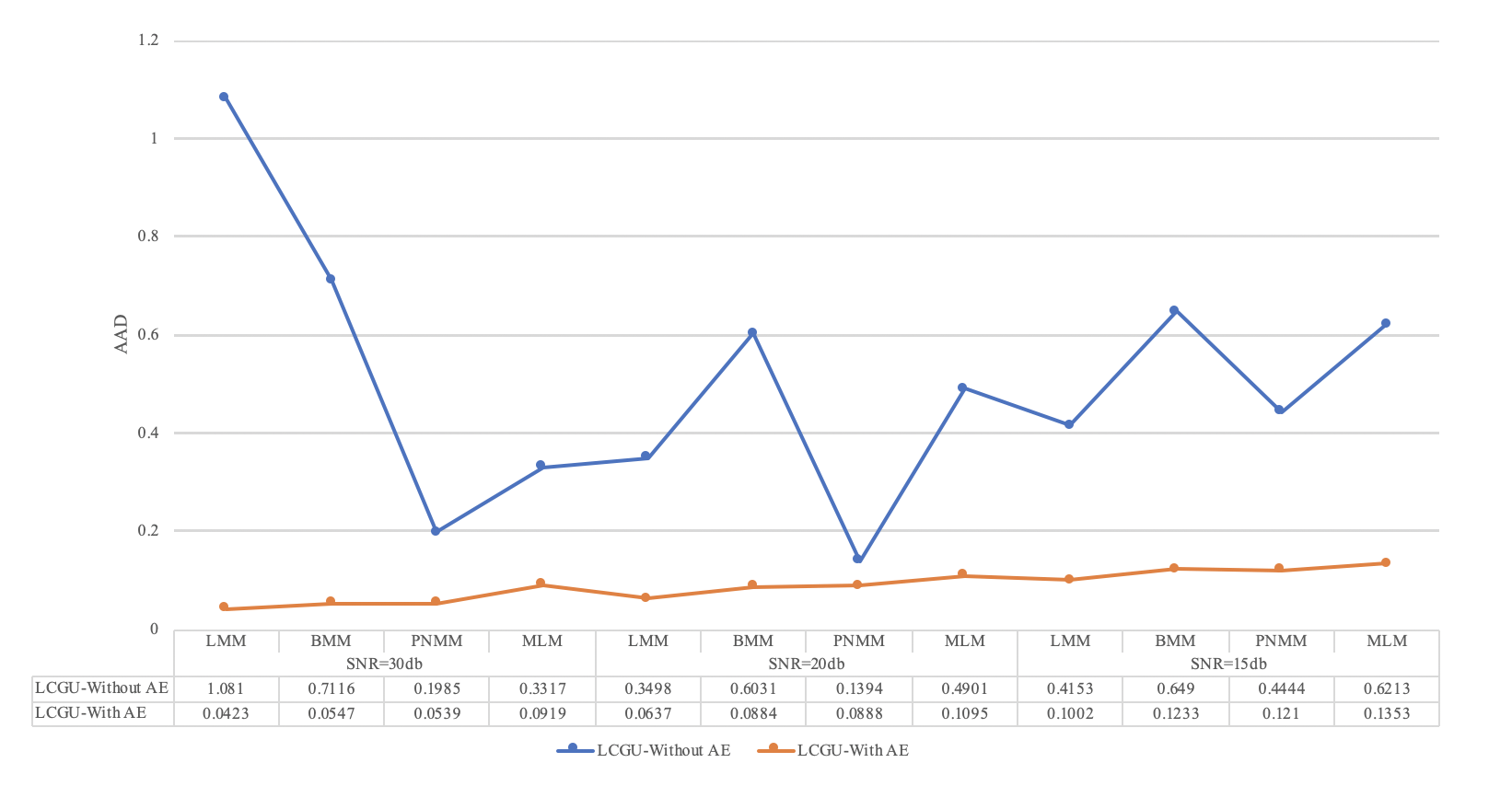}\\
  \caption{The effect of the semantic constraint between linear and nonlinear mixtures.}\label{fig:sup}
  \vspace{-3mm}
\end{figure}

\begin{figure}
  \centering
  \includegraphics[width=3.3in]{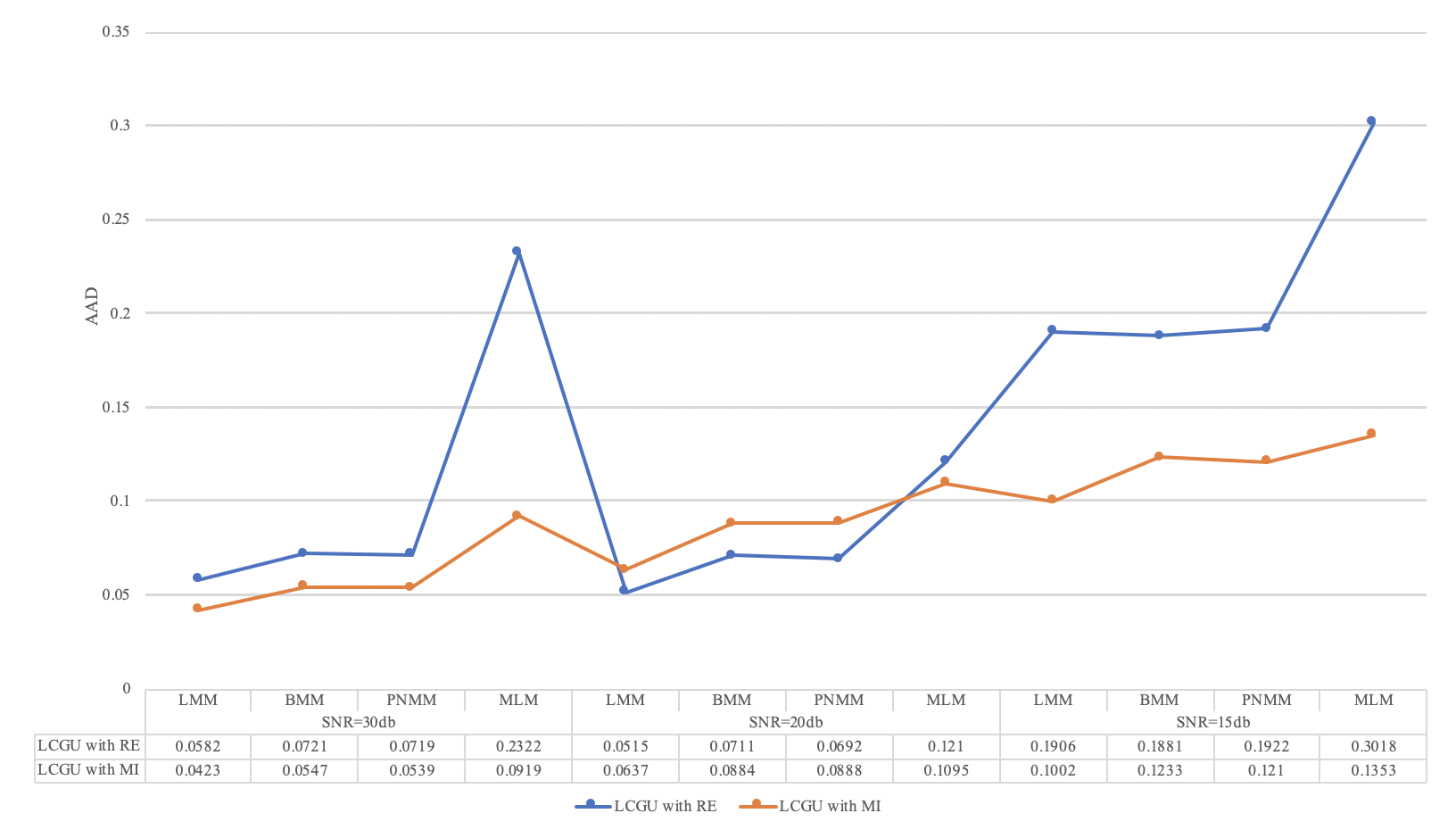}\label{fig:MI}\\
  \caption{The effect of the reconstruction loss versus semantic loss when incorporating the semantic constraint.}\label{fig:MI}
  \vspace{-3mm}
\end{figure}

\subsubsection{Ablation Analysis}
We further evaluate the contribution of the major components in the proposed \mn, including the bi-directional flow structure, the semantic constraint, and the mutual information loss, with three sets of experiments, and the results are shown in Figs.~\ref{fig:Bi}, \ref{fig:sup}, and \ref{fig:MI}, respectively. 

The comparison between the proposed method using the bi-directional flow and a uni-directional flow is shown in Fig.~\ref{fig:Bi}. The structure of the uni-directional flow was illustrated in single flow structure is shown at the green circle in Fig.\ref{fig:cyclegan}, \ie, $\mathcal{Y} \rightarrow \mathcal{A} \rightarrow \mathcal{Y}$. 
We can observe that the bi-directional structure can maintain a more stable performance on different types of mixture, as compared to the uni-directional structure. In addition, although the different SNR levels and mixture complexities affect the performance of both structures, the fluctuation in AAD is much smaller with the bi-directional structure, especially for datasets using 
complex mixing models, such as MLM, and datasets corrupted with lower SNR (\eg, SNR=15db). This indicates that training the network with a bi-directional structure is an effective way to improve the robustness and generalization capacity of the network.

According to the analysis in Sec.~\ref{sec:linear}, to preserve the semantic information of the derived abundance, we introduced the so-called semantic constraint. 
In this experiment, synthetic datasets with different SNRs are used to evaluate the effectiveness of this constraint, and the AAD of the results are shown in Fig.~\ref{fig:sup}. 
We compare the proposed \mn~structure with the direct usage of CycleGAN. 
From Fig.~\ref{fig:sup}, we can observe that the AAD's of the proposed \mn~on all synthetic datasets is significantly lower than those from the CycleGAN, which demonstrates that the semantic constraint that preserves the content structure of the abundance map is essential in improving the model-free unmixing performance.

Also in Sec.~\ref{sec:linear}, we described that to further preserve the semantic information of the estimated abundance, the mutual information (MI) loss $loss_3$ instead of the reconstruction loss between the linear and nonlinear mixtures is introduced to regularize the solution space. Figure~\ref{fig:MI} shows using these two different types of losses would affect the performance of the proposed \mn. 
The results demonstrate the superiority of the proposed method with MI $loss_3$, as compared to RMSE $loss_3$, where the AAD of the derived abundance from mutual information is smaller than that of the RMSE loss in most scenarios, especially when the data is contaminated with high-level noise, \eg, SNR=15db. This is because MI focuses on the global and semantic similarities between the linear mixtures and nonlinear mixtures rather than their local difference, which will be affected heavily by noise. This indicates that the proposed \mn~ with mutual information is more robust to noise. 

Based on above results and analysis, we can claim that the proposed bi-directional structure, semantic constraint, and the mutual information loss are able to improve the robustness and generalization capacity of the proposed \mn~effectively, paving the way to model-free unsupervised nonlinear unmixing.

\subsection{Results and Analysis of Real Data}
We further evaluate the proposed method on real datasets, \ie, the Urban and WDC hyperspectral images, with both quantitative and visual comparisons. 

The first group of experiments are conducted on the Urban dataset, and the results are presented in Table~\ref{tab:sad_urban}. Since the ground truth abundance is not available, the results are evaluated with the reconstruction error (RE) and SAD. The results in Table~\ref{tab:sad_urban} demonstrates that the proposed \mn~is able to reconstruct the image better with the derived abundance, as compared to the other unmixing methods. 

The estimated abundance map is also evaluated with visual inspection as shown in supplement. Because of the lacking of ground truth abundance, we download the high resolution image of the same Urban region from Google Earth, which can provide detailed information of objects, to facilitate the visual inspection of the estimated abundance. 
From abundance map, we can find that the objects consisting of different materials, \eg, Asphalt road in the first row and Roof in the fourth row, can be segmented better by the abundance estimated from the proposed \mn, which suggests that the proposed model-free method is more reliable than the other approaches on real data.  

The results of the WDC image shown in Table~\ref{tab:sad_wdc} confirm the findings in previous experiments. 
Similarly, the high resolution reference image covering the same WDC area is downloaded from Google Earth. Table~\ref{tab:sad_wdc} shows the quantitative comparison of different approaches using RE and SAD on the reconstructed spectra. Same observations can be made from this real dataset as in the Urban dataset. 
The abundance map, in the supplement, also supports the same conclusions drawn from the Urban image experiments.

\begin{table}
  \centering
  \caption{Comparison of Reconstruction Performance Using Different Unmixing Methods on the Urban Image }\label{tab:sad_urban}
  \vspace{2mm}
  \setlength{\tabcolsep}{1mm}{
  \begin{tabular}{c|c|c|c|c|c|c}
  \toprule
  \midrule
   {}&FCLS&PPNM&MLM&uDAS&NN-LM&\mn\\
    \midrule
    RE&0.0684&0.0661&0.0655&0.0685&0.0141&0.0134\\
    \midrule
    SAD&0.1499&0.1433&0.1467&0.0921&0.0793&0.0680\\
    \bottomrule
  \end{tabular}}
  \vspace{1mm}
\end{table}

\begin{table}
  \centering
  \caption{Comparison of Reconstruction Performance Using Different Unmixing Methods on the WDC Image }\label{tab:sad_wdc}
  \vspace{2mm}
  \setlength{\tabcolsep}{1mm}{
  \begin{tabular}{c|c|c|c|c|c|c}
  \toprule
  \midrule
   {}&FCLS&PPNM&MLM&uDAS&NN-LM&\mn\\
    \midrule
    RE&0.0309&0.0261&0.0166&0.0409&0.0263&0.0166\\
    \midrule
    SAD&0.0658&0.0605&0.0.0550&0.0606&0.0602&0.0488\\
    \bottomrule
  \end{tabular}}
  \vspace{-5mm}
\end{table}





\section{Conclusion}\label{conclude}
In this paper, we proposed a model-free hyperspectral nonlinear unmixing approach, \mn, through semanticly-constrained CycleGAN, where the solution space is regularized through both cycle consistency and semantic consistency using a bi-directional generative adversarial network flow. 
Extensive experiments on datasets constructed from different mixing models (both linear and nonlinear) demonstrated the superiority of the proposed \mn~over state-of-the-art,  
showing its strong generalization capacity. In our future work, we will exploit the nonlinear unmixing problem 
through unsupervised model-free network, where the endmembers are not known \textit{a-priori}.

{\small
\bibliographystyle{ieee_fullname}
\bibliography{egbib}

@inproceedings{gross2021multi,
  title={A multi-temporal hyperspectral target detection experiment: Evaluation of military setups},
  author={Gross, Wolfgang and Queck, Florian and V{\"o}gtli, Marius and Schreiner, Simon and Kuester, Jannick and B{\"o}hler, Jonas and Mispelhorn, Jonas and Kneub{\"u}hler, Mathias and Middelmann, Wolfgang},
  booktitle={Target and Background Signatures VII},
  volume={11865},
  pages={38--48},
  year={2021},
  organization={SPIE}
}

@inproceedings{vibhute2021hyperspectral,
  title={Hyperspectral Image Unmixing for Land Cover Classification},
  author={Vibhute, Amol D and Gaikwad, Sandeep V and Kale, Karbhari V and Mane, Arjun V},
  booktitle={2021 IEEE India Council International Subsections Conference (INDISCON)},
  pages={1--5},
  year={2021},
  organization={IEEE}
}

@inproceedings{guo2015hyperspectral,
  title={Hyperspectral image unmixing using autoencoder cascade},
  author={Guo, Rui and Wang, Wei and Qi, Hairong},
  booktitle={2015 7th Workshop on Hyperspectral Image and Signal Processing: Evolution in Remote Sensing (WHISPERS)},
  pages={1--4},
  year={2015},
  organization={IEEE}
}

@article{palsson2018hyperspectral,
  title={Hyperspectral unmixing using a neural network autoencoder},
  author={Palsson, Burkni and Sigurdsson, Jakob and Sveinsson, Johannes R and Ulfarsson, Magnus O},
  journal={IEEE Access},
  volume={6},
  pages={25646--25656},
  year={2018},
  publisher={IEEE}
}

@article{palsson2020convolutional,
  title={Convolutional autoencoder for spectral--spatial hyperspectral unmixing},
  author={Palsson, Burkni and Ulfarsson, Magnus O and Sveinsson, Johannes R},
  journal={IEEE Transactions on Geoscience and Remote Sensing},
  volume={59},
  number={1},
  pages={535--549},
  year={2020},
  publisher={IEEE}
}

@article{qu2018udas,
  title={uDAS: An untied denoising autoencoder with sparsity for spectral unmixing},
  author={Qu, Ying and Qi, Hairong},
  journal={IEEE Transactions on Geoscience and Remote Sensing},
  volume={57},
  number={3},
  pages={1698--1712},
  year={2018},
  publisher={IEEE}
}

@article{su2019daen,
  title={DAEN: Deep autoencoder networks for hyperspectral unmixing},
  author={Su, Yuanchao and Li, Jun and Plaza, Antonio and Marinoni, Andrea and Gamba, Paolo and Chakravortty, Somdatta},
  journal={IEEE Transactions on Geoscience and Remote Sensing},
  volume={57},
  number={7},
  pages={4309--4321},
  year={2019},
  publisher={IEEE}
}

@article{hinton2006reducing,
  title={Reducing the dimensionality of data with neural networks},
  author={Hinton, Geoffrey E and Salakhutdinov, Ruslan R},
  journal={Science},
  volume={313},
  number={5786},
  pages={504--507},
  year={2006},
  publisher={American Association for the Advancement of Science}
}

@article{hong2021endmember,
  title={Endmember-guided unmixing network (EGU-Net): A general deep learning framework for self-supervised hyperspectral unmixing},
  author={Hong, Danfeng and Gao, Lianru and Yao, Jing and Yokoya, Naoto and Chanussot, Jocelyn and Heiden, Uta and Zhang, Bing},
  journal={IEEE Transactions on Neural Networks and Learning Systems},
  volume={33},
  number={11},
  pages={6518--6531},
  year={2021},
  publisher={IEEE}
}

@article{wang2019nonlinear,
  title={Nonlinear unmixing of hyperspectral data via deep autoencoder networks},
  author={Wang, Mou and Zhao, Min and Chen, Jie and Rahardja, Susanto},
  journal={IEEE Geoscience and Remote Sensing Letters},
  volume={16},
  number={9},
  pages={1467--1471},
  year={2019},
  publisher={IEEE}
}

@article{zhao2021lstm,
  title={LSTM-DNN based autoencoder network for nonlinear hyperspectral image unmixing},
  author={Zhao, Min and Yan, Longbin and Chen, Jie},
  journal={IEEE Journal of Selected Topics in Signal Processing},
  volume={15},
  number={2},
  pages={295--309},
  year={2021},
  publisher={IEEE}
}

@article{zhao2021hyperspectral,
  title={Hyperspectral unmixing for additive nonlinear models with a 3-D-CNN autoencoder network},
  author={Zhao, Min and Wang, Mou and Chen, Jie and Rahardja, Susanto},
  journal={IEEE Transactions on Geoscience and Remote Sensing},
  volume={60},
  pages={1--15},
  year={2021},
  publisher={IEEE}
}

@article{koirala2019supervised,
  title={A supervised method for nonlinear hyperspectral unmixing},
  author={Koirala, Bikram and Khodadadzadeh, Mahdi and Contreras, Cecilia and Zahiri, Zohreh and Gloaguen, Richard and Scheunders, Paul},
  journal={Remote Sensing},
  volume={11},
  number={20},
  pages={2458},
  year={2019},
  publisher={MDPI}
}

@article{su2020deep,
  title={Deep autoencoders with multitask learning for bilinear hyperspectral unmixing},
  author={Su, Yuanchao and Xu, Xiang and Li, Jun and Qi, Hairong and Gamba, Paolo and Plaza, Antonio},
  journal={IEEE Transactions on Geoscience and Remote Sensing},
  volume={59},
  number={10},
  pages={8615--8629},
  year={2020},
  publisher={IEEE}
}

@article{gao2021cycu,
  title={CyCU-Net: Cycle-consistency unmixing network by learning cascaded autoencoders},
  author={Gao, Lianru and Han, Zhu and Hong, Danfeng and Zhang, Bing and Chanussot, Jocelyn},
  journal={IEEE Transactions on Geoscience and Remote Sensing},
  volume={60},
  pages={1--14},
  year={2021},
  publisher={IEEE}
}

@article{chen2013nonlinear,
  title={Nonlinear estimation of material abundances in hyperspectral images with l1-norm spatial regularization},
  author={Chen, Jie and Richard, C{\'e}dric and Honeine, Paul},
  journal={IEEE Transactions on Geoscience and Remote Sensing},
  volume={52},
  number={5},
  pages={2654--2665},
  year={2013},
  publisher={IEEE}
}

@inproceedings{swayze1993us,
  title={The US geological survey, digital spectral library: Version 1: 0.2 to 3.0 mum},
  author={Swayze, GA and Clark, RN and King, TVV and Gallagher, A and Calvin, WM},
  booktitle={Bulletin of the American astronomical society},
  volume={25},
  pages={1033},
  year={1993}
}

@article{tang2017integrating,
  title={Integrating spatial information in the normalized P-linear algorithm for nonlinear hyperspectral unmixing},
  author={Tang, Maofeng and Gao, Lianru and Marinoni, Andrea and Gamba, Paolo and Zhang, Bing},
  journal={IEEE Journal of Selected Topics in Applied Earth Observations and Remote Sensing},
  volume={11},
  number={4},
  pages={1179--1190},
  year={2017},
  publisher={IEEE}
}

@article{altmann2012supervised,
  title={Supervised nonlinear spectral unmixing using a postnonlinear mixing model for hyperspectral imagery},
  author={Altmann, Yoann and Halimi, Abderrahim and Dobigeon, Nicolas and Tourneret, Jean-Yves},
  journal={IEEE Transactions on Image Processing},
  volume={21},
  number={6},
  pages={3017--3025},
  year={2012},
  publisher={IEEE}
}

@article{halimi2011nonlinear,
  title={Nonlinear unmixing of hyperspectral images using a generalized bilinear model},
  author={Halimi, Abderrahim and Altmann, Yoann and Dobigeon, Nicolas and Tourneret, Jean-Yves},
  journal={IEEE Transactions on Geoscience and Remote Sensing},
  volume={49},
  number={11},
  pages={4153--4162},
  year={2011},
  publisher={IEEE}
}

@article{bioucas2012hyperspectral,
  title={Hyperspectral unmixing overview: Geometrical, statistical, and sparse regression-based approaches},
  author={Bioucas-Dias, Jos{\'e} M and Plaza, Antonio and Dobigeon, Nicolas and Parente, Mario and Du, Qian and Gader, Paul and Chanussot, Jocelyn},
  journal={IEEE journal of selected topics in applied earth observations and remote sensing},
  volume={5},
  number={2},
  pages={354--379},
  year={2012},
  publisher={IEEE}
}

@article{heylen2015multilinear,
  title={A multilinear mixing model for nonlinear spectral unmixing},
  author={Heylen, Rob and Scheunders, Paul},
  journal={IEEE transactions on geoscience and remote sensing},
  volume={54},
  number={1},
  pages={240--251},
  year={2015},
  publisher={IEEE}
}

@inproceedings{qu2018unsupervised,
  title={Unsupervised sparse dirichlet-net for hyperspectral image super-resolution},
  author={Qu, Ying and Qi, Hairong and Kwan, Chiman},
  booktitle={Proceedings of the IEEE conference on computer vision and pattern recognition},
  pages={2511--2520},
  year={2018}
}

@article{goodfellow2020generative,
  title={Generative adversarial networks},
  author={Goodfellow, Ian and Pouget-Abadie, Jean and Mirza, Mehdi and Xu, Bing and Warde-Farley, David and Ozair, Sherjil and Courville, Aaron and Bengio, Yoshua},
  journal={Communications of the ACM},
  volume={63},
  number={11},
  pages={139--144},
  year={2020},
  publisher={ACM New York, NY, USA}
}

@article{tang2018multiharmonic,
  title={Multiharmonic postnonlinear mixing model for hyperspectral nonlinear unmixing},
  author={Tang, Maofeng and Zhang, Bing and Marinoni, Andrea and Gao, Lianru and Gamba, Paolo},
  journal={IEEE Geoscience and Remote Sensing Letters},
  volume={15},
  number={11},
  pages={1765--1769},
  year={2018},
  publisher={IEEE}
}

@article{marinoni2015novel,
  title={A novel approach for efficient $ p $-linear hyperspectral unmixing},
  author={Marinoni, Andrea and Gamba, Paolo},
  journal={IEEE Journal of Selected Topics in Signal Processing},
  volume={9},
  number={6},
  pages={1156--1168},
  year={2015},
  publisher={IEEE}
}

@article{dobigeon2013nonlinear,
  title={Nonlinear unmixing of hyperspectral images: Models and algorithms},
  author={Dobigeon, Nicolas and Tourneret, Jean-Yves and Richard, C{\'e}dric and Bermudez, Jos{\'e} Carlos M and McLaughlin, Stephen and Hero, Alfred O},
  journal={IEEE Signal processing magazine},
  volume={31},
  number={1},
  pages={82--94},
  year={2013},
  publisher={IEEE}
}

@article{mustard1989photometric,
  title={Photometric phase functions of common geologic minerals and applications to quantitative analysis of mineral mixture reflectance spectra},
  author={Mustard, John F and Pieters, Carl{\'e} M},
  journal={Journal of Geophysical Research: Solid Earth},
  volume={94},
  number={B10},
  pages={13619--13634},
  year={1989},
  publisher={Wiley Online Library}
}

@book{hapke2012theory,
  title={Theory of reflectance and emittance spectroscopy},
  author={Hapke, Bruce},
  year={2012},
  publisher={Cambridge university press}
}

@article{nash1974spectral,
  title={Spectral reflectance systematics for mixtures of powdered hypersthene, labradorite, and ilmenite},
  author={Nash, DB and Conel, JE},
  journal={Journal of Geophysical Research},
  volume={79},
  number={11},
  pages={1615--1621},
  year={1974},
  publisher={Wiley Online Library}
}

@article{clark1984reflectance,
  title={Reflectance spectroscopy: Quantitative analysis techniques for remote sensing applications},
  author={Clark, Roger N and Roush, Ted L},
  journal={Journal of Geophysical Research: Solid Earth},
  volume={89},
  number={B7},
  pages={6329--6340},
  year={1984},
  publisher={Wiley Online Library}
}

@inproceedings{singer1979mars,
  title={Mars-large scale mixing of bright and dark surface materials and implications for analysis of spectral reflectance},
  author={Singer, Robert B and McCord, Thomas B},
  booktitle={Lunar and Planetary Science Conference Proceedings},
  volume={10},
  pages={1835--1848},
  year={1979}
}

@inproceedings{belghazi2018mutual,
  title={Mutual information neural estimation},
  author={Belghazi, Mohamed Ishmael and Baratin, Aristide and Rajeshwar, Sai and Ozair, Sherjil and Bengio, Yoshua and Courville, Aaron and Hjelm, Devon},
  booktitle={International conference on machine learning},
  pages={531--540},
  year={2018},
  organization={PMLR}
}
}

\end{document}